\documentclass[conference]{IEEEtran}

\IEEEoverridecommandlockouts
\usepackage{cite}
\usepackage{amsmath,amssymb,amsfonts}
\usepackage{algorithm}
\usepackage{algpseudocode}
\usepackage{graphicx}
\usepackage{hyperref}
\usepackage{textcomp}
\usepackage[table,xcdraw]{xcolor}
\usepackage{soul}
\usepackage{adjustbox}
\def\BibTeX{{\rm B\kern-.05em{\sc i\kern-.025em b}\kern-.08em
    T\kern-.1667em\lower.7ex\hbox{E}\kern-.125emX}}

\makeatletter
\def\ps@IEEEtitlepagestyle{%
	\def\@oddfoot{\mycopyrightnotice}%
	\def\@evenfoot{}%
}
\def\mycopyrightnotice{%
	{\hfill \footnotesize 978-1-6654-5705-7/23/\$31.00~\copyright 2023 IEEE\hfill}
}
\newcommand\notsotiny{\@setfontsize\notsotiny\@vipt\@viipt}
\makeatother

\begin{document}

\title{Towards Energy-Aware Federated Traffic Prediction for Cellular Networks

\thanks{This publication has been partially funded by the European Union's Horizon 2020 research and innovation programme under Grant Agreement No. 957406 (TERMINET), Grant Agreement No. 953775 (GREENEDGE) and the Grant CHIST-ERA-20-SICT-004 (SONATA) by PCI2021-122043-2A/AEI/10.13039/501100011033}}

\makeatletter
\newcommand{\linebreakand}{%
  \end{@IEEEauthorhalign}
  \hfill\mbox{}\par
  \mbox{}\hfill\begin{@IEEEauthorhalign}
}
\makeatother

\author{
    \IEEEauthorblockN{
        Vasileios Perifanis\IEEEauthorrefmark{1}, 
        Nikolaos Pavlidis\IEEEauthorrefmark{1},
        Selim F. Yilmaz\IEEEauthorrefmark{2},
        Francesc Wilhelmi\IEEEauthorrefmark{3},
        Elia Guerra\IEEEauthorrefmark{4}\\
        Marco Miozzo\IEEEauthorrefmark{4},
        Pavlos S. Efraimidis\IEEEauthorrefmark{1}, 
        Paolo Dini\IEEEauthorrefmark{4} and
        Remous-Aris Koutsiamanis\IEEEauthorrefmark{5}
    }
    \IEEEauthorblockA{
        \IEEEauthorrefmark{1}Department of Electrical and Computer Engineering, Democritus University of Thrace, Xanthi, Greece \\
        \IEEEauthorrefmark{2}Department of Electrical and Electronic Engineering, Imperial College London, London, United Kingdom  \\
        \IEEEauthorrefmark{3}Radio Systems Research, Nokia Bell Labs, Stuttgart, Germany \\
        \IEEEauthorrefmark{4}Sustainable Artificial Intelligence, Centre Tecnol\`ogic de Telecomunicacions de Catalunya (CTTC/CERCA), Barcelona, Spain \\
        \IEEEauthorrefmark{5}Department of Automation, Production and Computer Sciences, IMT Atlantique, Inria, LS2N, Nantes, France \\
    }
    \IEEEauthorblockA{e-mail: \{vperifan, npavlidi\}@ee.duth.gr}
}



\maketitle
\thispagestyle{plain}
\pagestyle{plain}

\begin{abstract}
Cellular traffic prediction is a crucial activity for optimizing networks in fifth-generation (5G) networks and beyond, as accurate forecasting is essential for intelligent network design, resource allocation and anomaly mitigation. Although machine learning (ML) is a promising approach to effectively predict network traffic, the centralization of massive data in a single data center raises issues regarding confidentiality, privacy and data transfer demands. To address these challenges, federated learning (FL) emerges as an appealing ML training framework which offers high accurate predictions through parallel distributed computations. However, the environmental impact of these methods is often overlooked, which calls into question their sustainability. In this paper, we address the trade-off between accuracy and energy consumption in FL by proposing a novel sustainability indicator that allows assessing the feasibility of ML models. Then, we comprehensively evaluate state-of-the-art deep learning (DL) architectures in a federated scenario using real-world measurements from base station (BS) sites in the area of Barcelona, Spain. Our findings indicate that larger ML models achieve marginally improved performance but have a significant environmental impact in terms of carbon footprint, which make them impractical for real-world applications.
\end{abstract}


\section{Introduction}

The advent of fifth-generation (5G) networks has brought forth a plethora of increasingly communication-dependent applications~\cite{Attaran20235G}, including autonomous driving~\cite{Aissioui20185G}, healthcare~\cite{Dananjayan20215G} and real-time recommender systems~\cite{Perifanis2022FedNCF}. To address the increasing complexity faced by 5G communications and beyond, network traffic forecasting emerges as an essential tool for proactively managing and operating networks.

Machine learning (ML) holds a great potential to undertake the network traffic forecasting task, as it may offer real-time and highly accurate predictions~\cite{Miozzo2021Distributed, zhang2018long}. More specifically, Deep Learning (DL) techniques such as Long Short-Term Memory (LSTM) networks~\cite{Nan2022FL} and transformers~\cite{Zeng2022AreTE, Zhang2023Crossformer} have shown remarkable performance in cellular traffic prediction. However, deploying such DL algorithms often face limitations in aggregating data from diverse sources due to regulatory restrictions, high bandwidth requirements and business competitiveness issues~\cite{Subramanya20215G,Piovesan2021Traffic, Perifanis20225g}. These issues are particularly stressed in traditional centralized ML settings, which struggle to cope with massively distributed data due to privacy concerns and communication overheads.

As a result, edge computing and distributed ML have garnered considerable attention in the recent years~\cite{Deng2020Edge}, as they improve privacy and address energy-related issues by reducing data movement and using hardware with limited resources~\cite{Ahvar2022Edge}. Among several distributed ML mechanisms, Federated Learning (FL) \cite{McMahan2017FedAvg} emerges as a popular solution for collaboratively training ML models without requiring raw data exchange. FL effectively tackles challenges related to multi-operator collaboration and multi-domain (geographical) problems within a single operator~\cite{Zhou2019Edge}, which makes it an appealing tool to realize traffic forecasting in future communications networks. Furthermore, FL holds the promise of enhanced accuracy and reduced environmental impact since it avoids heavy communication overheads and additional energy costs, such as cooling energy, incurred in big data centers~\cite{Savazzi2023FlFootpring}.

The foreseen benefits of FL are however threatened by the rapid increase in data volumes and the adoption of large-scale deep learning models, which demand substantial storage capacity and network bandwidth, while the accuracy improvements from these complex models often come at a significant environmental cost \cite{Wu2022SustainableAI, Strubell2019Energy, Meulemeester2023Sustainability}. Several studies also demonstrate that, despite the advances of large models with respect to their accuracy, they do not significantly surpass simpler models in the domain of time series forecasting~\cite{Zeng2022AreTE, Zhang2023Crossformer}.

In this paper, we assess the environmental impact of training DL models for network traffic forecasting in a federated setting. By examining the trade-off between accuracy and energy consumption, we aim to provide valuable insights and raise awareness regarding the environmental implications posed by the development and deployment of distributed AI technologies in communications systems. 

Our main contributions are summarized as follows:
\begin{enumerate}
    \item We introduce a generic FL framework, which we use to compare the state-of-the-art DL architectures for cellular traffic forecasting.
    \item We present a novel indicator to assess the sustainability of ML models, which we use to showcase the trade-off between energy consumption and predictive accuracy. While we employ the proposed indicator specifically for cellular traffic forecasting, its applicability can be extended to numerous other applications.
    \item We evaluate the performance of the considered DL models using real-world traffic measurements collected from cellular Base Stations (BSs), in the area of Barcelona (Spain) between 2018 and 2019.
\end{enumerate}

The rest of this paper is organized as follows. Section~\ref{related_work} presents the related literature on federated cellular traffic prediction and sustainability assessment of ML models. Section~\ref{method} outlines the problem statement, describes the models used and introduces the sustainability indicator. Section~\ref{experiments} presents and discusses the experimental results. Section~\ref{conclusion} summarizes the findings and provides final remarks.

\section{Related Work}\label{related_work}
\subsection{Federated Learning for Cellular Traffic Prediction}
The number of 5G connections is estimated to reach 5 billion ($10^9$), by 2030~\cite{GSMA2020Report}, hence traffic prediction will be of utmost importance for designing and optimizing next generation communications systems. The diverse and complex patterns of human mobility contribute to traffic variations among BSs, emphasizing the need for reliable predictive models. Traffic characteristics can exhibit significant changes between weekdays, weekends and social events~\cite{Bejarano2021Social}, making traffic forecasting and infrastructure planning challenging.

Recent advances leverage DL approaches to predict traffic demands~\cite{Gupta2021Traffic, Piovesan2021Traffic}. More recently, research efforts have shifted towards the decentralization of ML operations, offering improvements regarding scalability through decentralized ML model training frameworks such as FL. FL can potentially boost the collaboration among network operators withholding private data which they are reluctant to share, but which if used would lead to powerful and robust traffic predictors~\cite{Zhang2021FL}.

In~\cite{Zhang2021FL}, the authors designed a client-shifted FL algorithm with a dual aggregation scheme using call detail records (CDR) collected in Italy between 2013 and 2014~\cite{Barlacchi2015Dataset}. Using the same dataset, Nan et al.~\cite{Nan2022FL} trained a federated LSTM model using a regional aggregation algorithm. Similarly, in~\cite{Zhang2022FL}, the authors used the aforementioned dataset and employed a federated meta-learning approach. Subramanya et al.~\cite{Subramanya20215G} compared several models, including LSTM, CNN-LSTM and LSTM-LSTM for time-series forecasting in 5G networks using a private dataset from a commercial network operator. In~\cite{Perifanis20225g}, the authors presented several models for federated traffic prediction and demonstrated that advanced aggregation algorithms do not significantly outperform the FedAvg baseline~\cite{McMahan2017FedAvg}, owing to the influence of non-IID data in cellular traffic data.

In contrast to previous works, which focused on the now obsolete CDR data (SMS, voice calling, Internet) \cite{Zhang2021FL, Nan2022FL, Zhang2022FL}, in this paper, we use a more recent dataset that comprises real measurements from Long Term Evolution (LTE) BSs. This dataset contains contemporary information accounting for the current usage of cellular networks. Moreover, we build upon the models used in~\cite{Subramanya20215G} by incorporating state-of-the-art transformer-based models. Finally, we extend the work from~\cite{Perifanis20225g} by designing a sustainability indicator for federated settings that considers both the training and inference phases, both critical for the adoption of ML solutions in communications systems.

\subsection{Sustainability of Machine Learning}

Several studies have explored the sustainability of ML from various perspectives, offering insights and recommendations for reducing the environmental impact of ML algorithms~\cite{Wynsberghe2021Sustainable}. Wu et al.~\cite{Wu2022SustainableAI} investigated the carbon footprint throughout the entire life-cycle of ML, showing that carbon emissions primarily originate from the training and inference stages. In~\cite{Brownlee2021AccuracyEnergy}, the authors measured the energy consumption associated to the training and inference of multi-layer perceptron (MLP) models, showing that significant energy saving of up to 50\% could be achieved by reducing the number of hidden layers and units, with a minimal drop in accuracy ranging from 1-2\%. Additionally, Savazzi et al.~\cite{Savazzi2023FlFootpring} showed that carbon emission reduction can be achieved using FL instead of centralized ML. Lastly, Guerra et al.~\cite{Guerra2023Cost} compared the environmental impact of FL, gossip FL and blockchain-based FL, raising several open issues regarding the environmental aspects of training models using distributed approaches. 

In contrast to existing research on ML sustainability, we focus on the emissions of FL when applied to cellular traffic forecasting. Furthermore, we quantify the trade-off between ML model accuracy and emissions, especially with regard to large-sized models like transformers. Our research goes beyond MLPs~\cite{Brownlee2021AccuracyEnergy} and focuses on several DL models applied in a real-world scenario. Ultimately, our goal is to contribute to the promotion of sustainable AI practices~\cite{Schwartz2020GreenAI}.

\section{Methodology} 
\label{method}

In this section, we present the problem formulation for cellular traffic prediction and discuss the FL scenario. Additionally, we provide an overview of the ML models used.

\subsection{Problem Statement and FL Formulation}
We consider a cellular network with $N$ BSs connected to a common edge server. At every timestep $t$, each BS $k$ obtains a vector of $d$ measurements, denoted by $x_t^{(k)} \in \mathbb{R}^d$. At timestep $T$, each BS $k$ predicts its multivariate target measurements $y_T^{(k)} \in \mathbb{R}^{d'}$, where $d'$ denotes the number of measurements to be predicted, using a sliding window of last $W$ local measurements $X_{T-W:T-1}^{(k)} \in \mathbb{R}^{W \times d}$, where $X_{T-W:T-1}^{(k)} = \begin{bmatrix}
    x_{T-W}^{(k)} & x_{T-W+1}^{(k)} & \cdots & x_{T-1}^{(k)}
\end{bmatrix}$. A common neural network model $f(\cdot)$ is utilized for generating predictions, i.e., $\hat{y}_T^{(k)} = f( X_{T-W:T-1}^{(k)} )$, aiming at minimizing the prediction error while considering the energy consumption. The specific input and output values in the considered scenario are presented is Section \ref{dataset}.

We utilize two widely adopted metrics to quantify prediction error for time series forecasting: \textit{i)} the normalized root mean squared error ($\mathrm{NRMSE}$) and \textit{ii)} the mean absolute error ($\mathrm{MAE}$).
In our setting, given $m$ different target observation samples, $\mathrm{MAE}$ and $\mathrm{NRMSE}$ for BS $k$ are defined as follows:
\begin{equation}
    \mathrm{MAE^{(k)}} = \frac{1}{md'}\sum_{i=1}^{m} \lVert \hat{y}^{(k)}_{T+i}-y^{(k)}_{T+i} \rVert_1,
\end{equation}
\begin{equation}
    \mathrm{NRMSE^{(k)}} = \frac{1}{\overline{y}^{(k)}}\sqrt{\frac{\sum_{i=1}^{m} \lVert \hat{y}^{(k)}_{T+i}-y^{(k)}_{T+i} \rVert_2^2 }{md'}},
\end{equation}
where $\overline{y}^{(k)}= \frac{1}{md'}\sum_{i=1}^{m}y^{(k)}_{T+i}$. The goal is to minimize the average of $\mathrm{NRMSE^{(k)}}$ and $\mathrm{MAE^{(k)}}$, respectively, across all the BSs.

To develop a common model that can predict measurements at each BS while benefiting from the training data of all the BSs, we employ an FL-based training strategy whereby the server orchestrates the model training process. In each federated round, the server distributes the current global model to the edge devices. Each device feeds its local dataset to the ML pipeline and locally trains the received model for a number of local epochs. After local training, the updated model parameters are sent back to the server. The server aggregates the received model weights to create the new global model. The entire process repeats for multiple rounds until the global model converges.
\begin{figure}[t!]
\centerline{\includegraphics[width=\columnwidth]{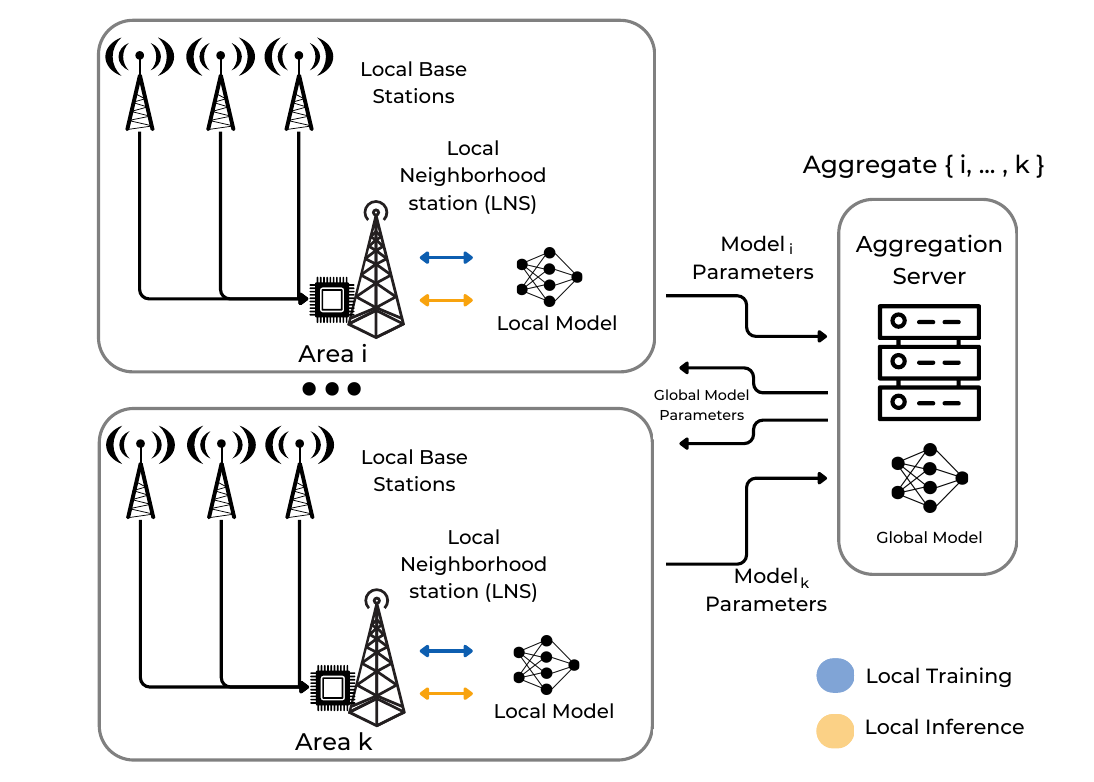}}
\caption{Federated learning-based traffic prediction in cellular networks.}
\label{fig:fl_architecture}
\end{figure}

In our learning scenario, we consider that each BS of a given area is associated with a Local Neighborhood Server (LNS) that collects data within its coverage. The LNS has sufficient resources to perform model training and orchestrate resource allocation at the monitored BSs. In this sense, each LNS serves as an FL node, communicating with the central server and performing local training and inference operations. Figure~\ref{fig:fl_architecture} provides an overview of the overall envisioned federated traffic prediction framework. Algorithm \ref{alg:alg1} summarizes the federated learning operations using the FedAvg algorithm \cite{McMahan2017FedAvg}. In our experimental study, the terms LNS and BS are equivalent and will be used interchangeably throughout the paper, as each LNS serves as the processing unit for each BS.

We develop a generic training methodology for accurate and energy-aware prediction of cellular traffic at each LNS. To achieve this, we assess different time series prediction models $f(\cdot)$ and monitor their associated energy consumption.
\begin{algorithm}[t]
\caption{Federated Learning for Cellular Traffic Forecasting with the FedAvg Algorithm.}\label{alg:alg1}
\textbf{Input:} Base stations $BSs = \{BS_1, BS_2, ..., BS_n\}$. $\mathcal{R}$ is the number federated rounds, $E$ is the number of local epochs, $B$ is the batch size, $\eta$ is the learning rate and $\nabla \mathcal{L}$ is the gradient optimization objective.\\
\textbf{Output:} Model weights $w$. 
\begin{algorithmic}[1]
\State Initialize $w_0$.
\For{each round $r = 1, 2, ..., \mathcal{R}$}
    \State $\{BS_r\} \leftarrow$ select round participants from $BSs$ at random without replacement.
  \State Transmit global model $w_{r-1}$ to LNSs that monitor each base station $k \in \{BS_r\}$
  \For{each base station $k \in \{BS_r\}$    \textbf{in parallel}}
    \State $w^k_{r} \leftarrow$ LocalTraining($k$, $w_{r-1}$)
  \EndFor
    \State $n_r \leftarrow \sum_{k \in \{BS_r\}}n_k$
    \State $w_{r+1} \leftarrow \sum_{k \in \{BS_r\}}\dfrac{n_k}{n_r}w_{r}^k$
\EndFor  
\Function{LocalTraining}{$k, w$} \algorithmiccomment{run on LNS monitoring BS $k$.}
    \State $\mathcal{B} \leftarrow$ split local dataset into batches of size $B$.
    \For{each local epoch $e=1, 2, ..., E$}
        \For{batch $b \in \mathcal{B}$}
            \State \hspace{0.5cm}$w \leftarrow w - \eta \nabla \mathcal{L}(w;b)$
        \EndFor
    \EndFor
\State \textbf{return} $w$ to server.
\EndFunction
\end{algorithmic}
\label{alg1}
\end{algorithm}

\subsection{Machine Learning Models}

To explore the trade-off between predictive accuracy and energy consumption, we adapt state-of-the-art DL models to the federated setting. We start with a vanilla LSTM as a baseline and then train encoder-decoder architectures as in~\cite{Subramanya20215G}. In addition, we explore additional models by integrating three transformer-based architectures, which have been widely adopted in various domains including time series forecasting~\cite{Zeng2022AreTE, Zhang2023Crossformer}. In particular, the following models are utilized:
\begin{enumerate}
    \item \textbf{LSTM:} The input series are fed into a single LSTM layer with 128 hidden units. The last output sequence by the LSTM layer is forwarded to a fully-connected layer of 128 units. The resulting hidden representation is then passed to the output layer to obtain the prediction.
    \item \textbf{CNN-LSTM:} The input series are passed through two one-dimensional convolutional layers with 32 channels and a kernel size of 1. The output of the CNN model is then processed by a single LSTM layer of 128 units and further processed through a fully-connected layer as for the previous model.%
    \item \textbf{LSTM-LSTM:} This architecture comprises two models, the LSTM encoder and the LSTM decoder. The input series are fed into the LSTM encoder with 128 units, which generates interpreted sequences that capture dependencies from the input domain and hidden representations. These encoded sequences are then forwarded to the LSTM decoder with 128 units, followed by a fully-connected layer of 128 units.
    \item \textbf{BasicTransformer:} This architecture contains the encoder part of a transformer~\cite{Vaswani2017Attention}. The input series are passed to a fully-connected layer of 128 units. The resulting hidden representation is then fed into 2 transformer blocks. Each transformer block applies multi-head attention using 8 heads, capturing complex dependencies throughout the input sequences. The output of the attention mechanism is normalized and further forwarded to a feed-forward neural network (FFNN) with 2 hidden layers, each comprising 128 units. Finally, the output is normalized and a fully-connected layer outputs a prediction.
    \item \textbf{Transformer:} This architecture extends the previous model by including a decoder sub-model. In this case, the output of the encoder and specifically, the last hidden representation, is fed into a fully-connected layer with 128 units. After that, 2 transformer blocks apply multi-head attention to the sum of the hidden representation obtained from the fully-connected layer and the entire output of the encoder. Transformer blocks are identical to the ones from the BasicTransformer. Finally, a fully-connected layer takes the last hidden representation obtained after applying multi-head attention and generates a prediction.
    \item \textbf{Transformer-LSTM:} This architecture enhances the previous Transformer model by including an additional LSTM layer. More specifically, the input series are processed by the encoder, which output is forwarded to an LSTM layer of 128 units. The last output sequence from the LSTM layer, along with the entire encoder’s output, are fed to the decoder, which generates predictions. 
\end{enumerate}

\subsection{Attention Mechanism}
Given the ability of attention mechanisms to focus on specific parts of features and subsequently lead to higher predictive accuracy \cite{Vaswani2017Attention, Zeng2022AreTE, Zhang2023Crossformer}, we also integrate an attention mechanism to models \emph{1)} to \emph{3)}. Note that transformer-based models \emph{4)} to \emph{6)} directly utilize self-attention mechanism, so additional attention is not required. More precisely, in the LSTM model, attention is applied to the output sequences obtained from the LSTM, i.e., before the final fully-connected layer. In the CNN-LSTM model, the attention is integrated after LSTM's operation. Lastly, in the LSTM-LSTM model, attention is incorporated into the encoder's output.

The attention mechanism considered in this work takes the final hidden states generated by an LSTM layer and repeats them for the specified window ($W=10$ in our case). The repeated hidden states and the LSTM's output sequence are concatenated to the last dimension and then forwarded to a fully-connected layer, followed by a \emph{tanh} activation. The resulting transformation of the hidden states undergoes a dot product operation with a learnable vector $v$. Finally, \emph{softmax} is applied to normalize the attention weights, which represent the importance of hidden features in the LSTM’s output sequence. The entire process can be summarized as
\begin{equation}
    \text{Attention($h$, $out$)} = \text{softmax}\left(v \cdot e\right),
\end{equation}
where
\begin{equation}
    e = \text{tanh}\left(
                \text{FC}\left(
                    \text{Concat}\left(\text{Repeat}\left(\text{$h$, timesteps}\right)\text{, }out\right)
                \right)
            \right),
\end{equation}
where $h$ and $out$ represent the LSTM's outputs, $v$ is a learnable weight vector, FC denotes a fully-connected layer and $\cdot$ represents matrix multiplication. In the remainder of this paper, models with the integrated attention mechanism are denoted with the suffix `-A', e.g., LSTM with attention is denoted as LSTM-A.

\subsection{Sustainability Indicator} \label{sust_indicator}

The absence of standardized indicators for evaluating ML sustainability poses challenges to performing fair comparisons among different algorithms. To address this issue, we propose a novel metric that considers both predictive accuracy and environmental impact. More specifically, we consider the following aspects: 
\begin{enumerate}
    \item Accuracy on unseen data, measured by the prediction error.
    \item Computational efficiency with respect to the energy consumed in Watt hours (Wh).
    \item Communication efficiency, quantified by the data size to be transmitted in kilobytes (kB).
\end{enumerate}
We select these aspects since accuracy indicates the training and inference reliability; computational efficiency is associated with energy consumption and environmental implications; communication efficiency relates to throughput and bandwidth requirements. These factors are crucial for assessing the sustainability of FL applications.

The sustainability indicator, denoted as $S$, provides a comprehensive evaluation of a model's sustainability throughout the training and inference phases. The formula for calculating $S$ is as follows:
\begin{equation}\label{eq:sustainability_formula}
    S = S_\text{Tr} \times S_\text{Inf},
\end{equation}
where $S_\text{Tr}$ and $S_\text{Inf}$ represent the attained trade-off between accuracy and energy consumption during the \textit{training} and \textit{inference} phases, respectively. The indicator for training can be calculated using the following equation:
\begin{equation}\label{eq:strain}
    S_\text{Tr} = (1 + E_\text{Val})^{\alpha} \times (1 + C_\text{Tr})^{\beta} \times (1 + DS)^{\gamma},
\end{equation}
where $E_\text{Val}$ is the validation error (in this work, we consider MAE), $C_\text{Tr}$ represents the total energy consumed for model training in Wh and $DS$ is the data size to be transmitted to the central server in kB. Under the FL setting, $DS$ represents the model size that is transmitted per federated round, while in a centralized scenario, it represents the raw dataset size. Note that, in this work, we solely focus on FL. The exponents denote the importance of each value, with $\alpha + \beta + \gamma = 1$. It should be clarified that a simple weighted average was not used due to the lack of normalization across the different scales, which could lead to misleading outcomes. As for the rationale behind $S_{Tr}$, a lower value indicates better computational and communication efficiency relative to accuracy. An ideal model has $E_\text{Val} = C_\text{Tr} = DS = 0$ and $S_\text{Tr} = 1$.

During inference, the following formula is utilized:
\begin{equation}\label{eq:sinference}
S_\text{Inf} = (1 + E_\text{Test})^{\alpha'} \times (1 + C_\text{Inf})^{\beta'}
\end{equation}
where $E_\text{Test}$ is the error observed in unseen test data (e.g., based on MAE) and $C_\text{Inf}$ is the energy consumed for predictions. The communication cost during the inference phase is not considered as each client holds its own model locally. Since the number of times that an FL client employs the model for generating predictions is specific to the implementation, we measure these values in terms of predictions per 1.000 samples. Note that, in the considered scenario, each BS uses the model every two minutes, resulting in 720 predictions per day. In this service, the operator is interested in knowing the predictions of the model's output variables to implement its network management optimization solutions. Similar to $S_\text{T}$, $\alpha' + \beta' = 1$ and, the lower the value of $S_\text{Inf}$, the better the trade-off between computational efficiency and predictive accuracy.

\section{Results}
\label{experiments}
In this section, we outline the experimental setup and present the results focusing on the predictive accuracy and energy consumption. Finally, we delve into the sustainability aspects of the considered models, examining their viability through the sustainability indicator proposed in this paper.\footnote{The code is available at \url{https://github.com/vperifan/federated-Time-Series-Forecasting/}.}

\subsection{Dataset and Experimental Details}
\label{dataset}

\begin{table*}[ht!]
\notsotiny
\caption{Dataset Statistics}
\begin{adjustbox}{width=\textwidth,center}
\begin{tabular}{lcccccccccccc}
\hline
Area    &   & Down    & Up    & RNTI Count  & MCS Down       & MCS Down Var  & MCS Up  & MCS UP Var & RB Down      & RB Down Var  & RB Up        & RB Up Var   \\
& & ($\times 10^9$) & ($\times 10^9$) & ($\times 10^4$) & ($\times 10^1$) & ($\times 10^2$) & ($\times 10^1$) & ($\times 10^2$) & ($\times 10^{-1}$) & ($\times 10^{-7}$) & ($\times 10^{-1}$) & ($\times 10^{-7}$)\\
\hline
ElBorn  &  Min & 0.005   & 0.0        &  0.035       & 0.193  & 0.337  & 0.0     & 0.0       & 0.009  & 0.182  &  0.0         & $0.0$          \\
        & Max  &  1.887 &  0.673 &  4.373      & 1.653  &  1.009  &  3.100     &  2.410    &  6.706  &  1.581  &  2.236  &  0.681   \\
\hline
LesCorts & Min &  0.0         &  0.0        &  0.0         &  0.0                &  0.0                &  0.0      &  0.0       &  0.0                 &  0.0                      &  0.0               &  0.0                \\
        & Max  &  0.297  &  1.058 &  2.135      &  1.616 &  0.993   &  3.100     &  1.477  &  1.241  &  0.522  &  2.314         &  0.940  \\
\hline
PobleSec  & Min &  0.008   &  0.0        &  0.056        &  0.086  &  0.152  &  0.0      &  0.0       &  0.015  &  0.231  &  0.0        &  0.0              \\
        & Max  &  2.286 &  0.625 &  1.619      &  1.619 &  0.969   &  3.050     &  1.823    &  7.398              &  1.801  &  4.458 &  1.026  \\
\hline
\end{tabular}
\end{adjustbox}
\label{tab:dataset_stats}
\end{table*}

The datasets were collected from three locations in Barcelona, Spain, representing different zones: touristic, entertainment and residential areas. The datasets ensure user anonymity and offer accurate information about the network utilization, allowing the extraction of detailed traces from individual communications. The datasets have been pre-processed to include aggregated statistics for every two-minute interval. Table~\ref{tab:dataset_stats} reports the minimum and maximum values for each location and type of measurement. In particular, the three locations are treated as distinct nodes under FL and each site has the following characteristics: 
\begin{itemize}
    \item \textbf{ElBorn:} 5.421 samples, collected from 2018-03-28 15:56:00 to 2018-04-04 22:36:00.
    \item \textbf{LesCorts:} 8.615 samples, collected from 2019-01-12 17:12:00 to 2019-01-24 16:20:00.
    \item \textbf{PobleSec:} 19.909 samples,  samples, collected from 2018-02-05 23:40:00 to 2018-03-05 15:16:00.
\end{itemize}

Following the analysis by \cite{Perifanis20225g}, the distributions and the number of observation significantly vary among these localities, resulting in non-IID data distribution. For the experimental evaluation, we use the standard 60/20/20 train, validation and test split per base station.

For each site, the eleven features given in Table \ref{tab:dataset_stats} are used as input to the ML models with a window of $W=10$. The goal is to predict the next timestep's five features: \textit{uplink} and \textit{downlink} traffic (Down and Up), the \textit{radio network temporary identifiers} (RNTI Count) and the \textit{resource blocks} for \textit{downlink} and \textit{uplink} (RB Down and RB Up). We employ 50 federated rounds with 3 local epochs per site, optimizing the MAE for the five target values. The energy consumption per considered model is measured using CodeCarbon,\footnote{\url{https://github.com/mlco2/codecarbon}} a tool that monitors the energy consumed either in GPU or CPU during the training. The experiments were conducted on a workstation running Ubuntu 22.04 with 64 GB memory and an Intel Xeon 4210R CPU and RTX A6000 GPU.

\begin{table*}[htbp]
    \caption{Results on Traffic Forecasting on Validation/Test sets and Model Sustainability.}
    \begin{adjustbox}{width=\textwidth,center}
    \centering
    \begin{tabular}{ll|cc|cccccc}
    \hline
     \textbf{Model} & \textbf{Area}  &   \textbf{NRMSE} &     \textbf{MAE} ($\times 10^6$) & $\mathbf{E_\text{Tr}}$ (Wh) & $\mathbf{E_\text{Inf}}$ (Wh) & \textbf{Size} (KB) & $\mathbf{S_\text{Tr}} (\times 10^3)$ & $\mathbf{S_\text{Inf}}  (\times 10^3)$ & $\mathbf{S} (\times 10^7)$ \\
    \hline
    LSTM & ElBorn       & 0.3925 / 0.8909 & 3.5868 / 8.9698\\
                & LesCorts   & 0.3142 / 0.2702 & 2.9398 / 3.2382  & 10.2346 & 0.0350 & 292.9 & 2.3050 & 2.7564 & 0.6353 \\
                & PobleSec     & 1.1091 / 0.8597 & 7.5419 / 9.8139\\
    \hline
    LSTM-A & ElBorn        & 0.3946 / 0.8619 & 3.5857 / 8.9025 \\
                & LesCorts    & 0.3026 / 0.2599 & 2.9501  / 3.2111 & 14.2425 & 0.0437 & 425.7 & 2.8854 & 2.7622 & 0.7970\\
                & PobleSec       & 1.1570 / 0.8488 & 7.5707 / 9.8169  \\
    \hline
    CNN-LSTM & ElBorn        & 0.4198 / 0.8967 & 4.0138 / 9.8041\\
                & LesCorts        & 0.3298 / 0.2775 & 3.0930 / 3.2873 & 12.8932 & 0.0439 & 342.5 & 2.6440 & 2.8290 & 0.7480\\
                & PobleSec           & 1.1498 / 0.8450 &  7.6466 / 9.9082\\
    \hline
    CNN-LSTM-A & ElBorn           & 0.4342 / 0.8814 & 4.0801 / 9.7783 \\
                & LesCorts   & 0.3194 / 0.2694 & 3.1196 / 3.2618 & 15.8389 & 0.0501 & 475.3 & 3.1426 & 2.8300 & 0.8894\\
                & PobleSec          &  1.1748 / 0.8450 & 7.6173 / 9.8418\\
    \hline
    LSTM-LSTM & ElBorn            & 0.4260 / 0.8965 & 3.6611 / 9.2160\\
                & LesCorts    & 0.3272 / 0.2822 & 3.0237 / 3.2961 & 13.1634 & 0.0450 & 821.9 & 3.5211 & 2.7941 & 0.9838\\
                & PobleSec           & 1.1415 / 0.8543 & 7.7067 / 9.9019 \\
    \hline
    LSTM-LSTM-A & ElBorn          &  0.4062 / 0.8922 & 3.5585 / 8.9310\\
                & LesCorts    & 0.3187 / 0.2726 & 2.9646 / 3.2280 & 16.3795 & 0.0577 & 954.8 & 3.9356 & 2.7843 & 1.0958\\
                & PobleSec          & 1.1294 / 0.8649 & 7.6249 / 9.8291 \\
    \hline
    BasicTransformer & ElBorn        &  0.4147 / 0.8372 & 3.7374 / 8.8270\\
                & LesCorts  & 0.4054 / 0.3380 & 3.1668 / 3.3568 & 25.2506 & 0.0851 & 814.9 & 4.3251 & 2.8149 & 1.2175\\
                & PobleSec      &  1.2533 / 0.8356 & 7.7016 / 9.7230  \\  
    \hline
    Transformer & ElBorn     &  0.4071 / 0.8328 & 3.5790 / 8.7340\\
                & LesCorts   & 0.3611 / 0.3041 & 3.0631 / 3.2598 & 39.5148 & 0.1233 & 1628.3 & 6.2182 & 2.8443 & 1.7687\\
                & PobleSec         &  1.2355 / 0.8299 & 7.5964 / 9.6134\\
    \hline
    Trasformer-LSTM & ElBorn    &  0.4045 / 0.8321 & 3.6243 / 8.7798\\
                & LesCorts      & 0.3357 / 0.2835 & 3.0506 / 3.2620 & 42.6441 & 0.1671 & 2217.3 & 7.0643 & 2.9061 & 2.0530\\
                & PobleSec       & 1.2490 / 0.8338 & 7.6149 / 9.6667 \\
    \hline
    \end{tabular}
    \end{adjustbox}
    \label{tab:valtest_results}
\end{table*}

\subsection{Forecasting Error}

The results per area considering the validation and test NRMSE and MAE of the forecasting task are presented in Table \ref{tab:valtest_results}. The following observations are made per BS:
\begin{itemize}
    \item \textbf{ElBorn:} In terms of validation and test MAE, LSTM-LSTM-A and Transformer are the top performing models. Although vanilla LSTM demonstrates the lowest NRMSE on the validation set, Transformer-LSTM performs the best on test NRMSE.
    \item \textbf{LesCorts:} Interestingly, simpler models yield better results for this particular BS. The LSTM model achieves the highest score considering the validation MAE, while the addition of attention mechanism enables the LSTM-A model to achieve the best scores on both the test MAE, validation NRMSE and test NRMSE.
    \item \textbf{PobleSec:} A significant distinction is observed between the validation and test scores for this BS. In terms of validation MAE and NRMSE, the LSTM model demonstrates the best performance. However, the Transformer model shows superior predictive accuracy when considering the test NRMSE and MAE.
\end{itemize}

Regarding the federated model that achieves the lowest average errors, its performance depends on whether we consider the validation or test set. On the validation set, the vanilla LSTM model shows slightly lower errors than
state-of-the-art Transformers, displaying the best accuracy.
For instance, the LSTM obtains 1.2\% more averaged validation MAE than Transformer. However, on the unseen test set, both the Transformer and Transformer-LSTM models demonstrate the lowest error outperforming the vanilla LSTM by 2\% and 1.5\%, respectively. The higher error on the test set for models such as the LSTM is attributed to the presence of overfitting, which affects their generalization capability. It is worth noting that there is no single model that performs best across all areas and metrics, indicating that the optimal model depends on the target area and the specific metric that is prioritized.

\begin{figure}[htbp]
\centerline{\includegraphics[width=0.85\columnwidth]{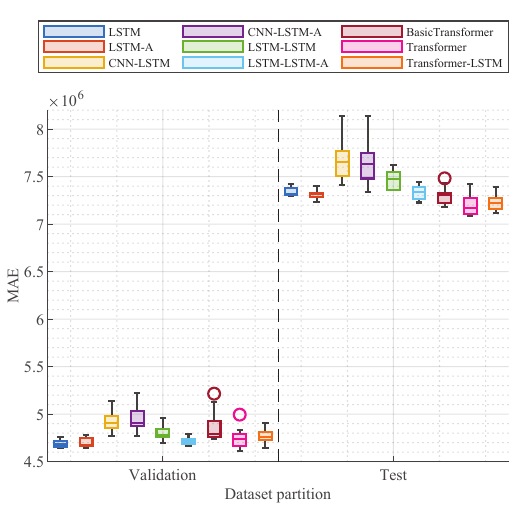}}
\caption{MAE per model on the validation and test sets.}
\label{fig:multi_results}
\end{figure}

In Fig. \ref{fig:multi_results}, we show the averaged MAE (across all areas) on the validation and test sets per model. As mentioned earlier, LSTM performs the best on the validation set, while Transformer and Transformer-LSTM show superior accuracy on the test set. These observations align with related works on time series forecasting, demonstrating that advanced models may not systematically outperform simpler ones \cite{Zeng2022AreTE, Zhang2023Crossformer}.

\subsection{Energy and Communication Costs}

In ML pipelines, energy is consumed during both the training and inference phases \cite{Wu2022SustainableAI}. In centralized settings, the training phase is typically performed once, while inference is repeated multiple times, when a client queries the trained model for predictions. In edge computing scenarios, frequent training is also needed due to real-time data collection, posing environmental constraints.

Communication costs are also relevant in both centralized and federated scenarios due to data collection and exchange of model weights. However, centralized scenarios involve additional costs such as cooling energy, while FL mitigates environmental costs by only transferring a small amount of data per federated round. This offers an advantage to FL compared to centralized learning regarding network throughput and latency, especially in large-scale scenarios. It also implies the assumption that, when transmitting larger models in the FL scenario, a greater environmental impact is anticipated. 

The model sizes that define the amount of data that needs to be transmitted per model broadcast, which quantifies the communication costs in an FL scenario, are presented in Table~\ref{tab:valtest_results} (in column \emph{Size}). As expected, integrating attention mechanisms leads to larger model sizes. Simpler models such as LSTM and CNN-LSTM have the smallest sizes, while Transformer-based models, exhibit much higher sizes. For instance, the Transformer model \cite{Vaswani2017Attention} is 5.6 times larger than the vanilla LSTM model. However, even the largest model (the Transformer-LSTM at 2217.3 KB) is still small enough in absolute terms to allow the FL methods to be usable in our scenario, since transferring this amount of data per BS per 120 seconds should be insignificant in comparison to the link capacities involved.

\begin{figure}[t!]
\centerline{\includegraphics[width=\columnwidth]{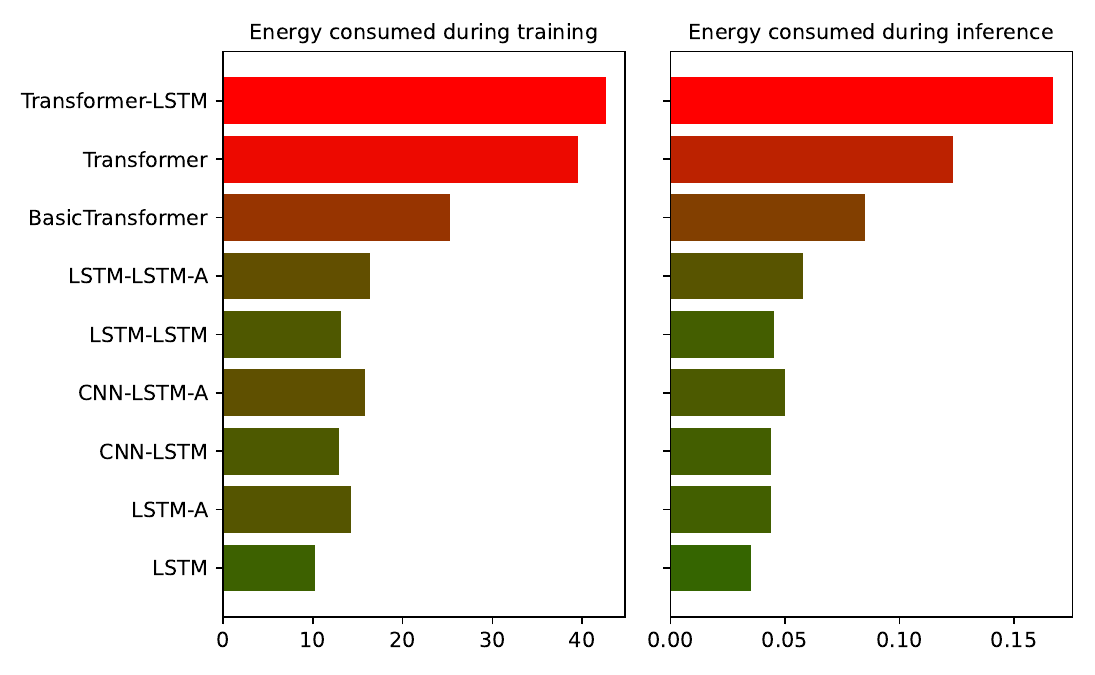}}
\caption{Total energy consumed (Wh) per model over 50 federated rounds. Training (left) and inference of 1.000 samples (right) costs are included.}
\label{fig:energy_consumption}
\end{figure}

In addition to the communication costs, we quantify the energy consumed during training and inference, which is depicted in Fig.~\ref{fig:energy_consumption} (also $E_\text{Tr}$ and $E_\text{Inf}$, respectively, in Table~\ref{tab:valtest_results}), using the CodeCarbon library. These values illustrate the total energy consumption (in Wh) per ML model during the federated training and inference phases. Specifically, for the inference phase, we consider the energy consumed for making 1.000 predictions. This quantification shows that more complex models like Transformers entail a higher energy cost. In particular, the BasicTransformer model consumes more than twice of the energy than the vanilla LSTM during the training phase, while the Transformer and Transformer-LSTM models have approximately five and seven times higher energy consumption, respectively. The CNN-LSTM and LSTM-LSTM models show significantly lower environmental impact compared to Transformers, but their energy is higher than the vanilla LSTM. It is worth noting that the inclusion of attention mechanisms increases the model complexity, size and energy consumption. Regarding the energy consumed during inference, we observe a similar behavior, where larger models consume significantly more energy compared to simpler ones. Overall, we conclude that more complex model architectures result in higher energy consumption during both the training and the inference phases, which can have a significant CO$_2$ footprint and overall environmental impact.

\begin{figure}[t!]
\centerline{\includegraphics[width=0.6301\columnwidth]{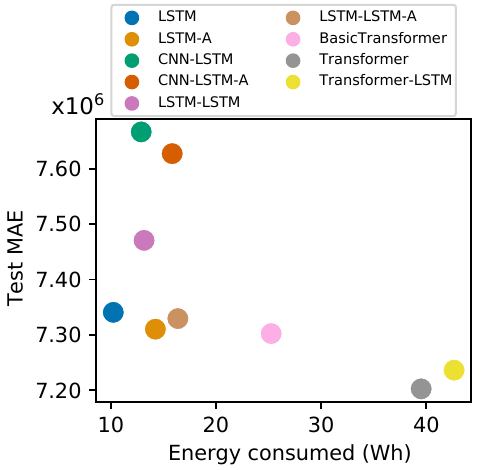}}
\caption{Trade-off between the test MAE and energy consumption (Wh) of each considered model.}
\label{fig:accuracy_energy}
\end{figure}

\subsection{Sustainability Evaluation}

This section presents an evaluation of the considered models using the sustainability indicator proposed in Section \ref{sust_indicator}. The formula in Eq.~\eqref{eq:sustainability_formula} takes into account the energy consumption during both the training and inference phases as well as the additional communication costs incurred during the exchange of model parameters between the server and FL nodes. The resulting values for each model under consideration are presented in Table \ref{tab:valtest_results} ($S_\text{Tr}, S_\text{Inf}$ and $S$). It is important to note that all the factors in Eq.~\eqref{eq:strain} and \eqref{eq:sinference} are equally weighted.

The resulting sustainability values reveal that more complex models such as those based on encoder-decoder or Transformer architectures exhibit poorer performance in terms of sustainability, both during training and inference. This can be attributed to their larger model sizes and higher energy consumption. Although these models demonstrate slightly better results in terms of prediction error, such a marginal advantage does not outweigh the significantly better results achieved by much simpler models in terms of sustainability. For instance, the resulting $S$ of the Transformer model is about 3, 2.5 and 2 times higher than the vanilla LSTM, CNN-LSTM and LSTM-LSTM models, respectively.

Figure \ref{fig:accuracy_energy} illustrates the trade-off between test MAE and energy consumption (in Wh) during the training phase of each respective model. This figure demonstrates that, although larger models such as Transformers lead to lower test MAE than simpler models such as the vanilla LSTM, the required energy for training them is the highest among models, making them less suitable when energy consumption is considered as an aspect for evaluating the overall performance.

\section{Conclusion}
\label{conclusion}

In this paper, we have investigated the sustainability and predictive performance of state-of-the-art DL models for federated cellular traffic forecasting. We have introduced a novel sustainability indicator for evaluating energy consumption with respect to accuracy, which enables convenient comparisons across various ML models in different experimental scenarios. We have shown that increasingly large and complex models provide very limited accuracy gains but have an enormous associated increase in energy consumption compared to simpler models. In the future, we aim to study the convergence speed of different models and extend the introduced sustainability indicator to capture aspects such as robustness. To demonstrate the generalization and scalability of federated learning, we will also evaluate an extensive and diverse set of clients and datasets. Finally, we will explore the trade-off between model selection and accuracy for each federated client and apply regularization techniques to improve model robustness.

\bibliographystyle{IEEEtran}
\bibliography{main}



\end{document}